\documentclass[journal]{IEEEtran}

\usepackage{epsfig,epstopdf}
\usepackage{cite}
\usepackage{graphicx}
\usepackage[cmex10]{amsmath}
\usepackage[ruled, lined, linesnumbered, commentsnumbered, longend]{algorithm2e}
\usepackage{xcolor}
\usepackage{multirow}

\usepackage{array}
\usepackage{amsfonts}
\graphicspath{ {./figs/} }

\usepackage{etoolbox}
\makeatletter
\patchcmd{\@makecaption}
{\scshape}
{}
{}
{}
\makeatletter
\patchcmd{\@makecaption}
{\\}
{.\ }
{}
{}
\makeatother
\def\tablename{Table}

\usepackage{color, colortbl}
\definecolor{LightCyan}{rgb}{0.89,1,1}
\definecolor{DarkCyan}{rgb}{0.5,1,1}

\usepackage{cellspace}
\begin{document}
%
\title{Constructing Accurate and Efficient Deep Spiking Neural Networks with Double-threshold and Augmented Schemes}
%
%
%

\author{
	Qiang~Yu,
	Chenxiang~Ma, 
	Shiming~Song, 
	Gaoyan~Zhang,\\
	Jianwu~Dang, 
	Kay~Chen~Tan,~\IEEEmembership{Fellow,~IEEE} 


\thanks{Q.~Yu, C.~Ma, S.~Song, G.~Zhang and J.~Dang are with Tianjin Key Laboratory of Cognitive Computing and Application, College of Intelligence and Computing, Tianjin University, Tianjin, China.}

\thanks{J.~Dang is also with Japan Advenced Institute of Science and Technology, Japan.}

\thanks{K.C.~Tan is with the Department of Computer Science, City University of Hong Kong, Hong Kong.}

\thanks{Corresponding author: Q.~Yu (e-mail: yuqiang@tju.edu.cn).}

}

%
%

\markboth{}%
{Shell \MakeLowercase{\textit{et al.}}: Bare Demo of IEEEtran.cls for Journals}
%



\maketitle

\begin{abstract}
Spiking neural networks (SNNs) are considered as a potential candidate to overcome current challenges such as the high-power consumption encountered by artificial neural networks (ANNs), however there is still a gap between them with respect to the recognition accuracy on practical tasks. A conversion strategy was thus introduced recently to bridge this gap by mapping a trained ANN to an SNN. However, it is still unclear that to what extent this obtained SNN can benefit both the accuracy advantage from ANN and high efficiency from the spike-based paradigm of computation. In this paper, we propose two new conversion methods, namely TerMapping and AugMapping. The TerMapping is a straightforward extension of a typical threshold-balancing method with a double-threshold scheme, while the AugMapping additionally incorporates a new scheme of augmented spike that employs a spike coefficient to carry the number of typical all-or-nothing spikes occurring at a time step. We examine the performance of our methods based on MNIST, Fashion-MNIST and CIFAR10 datasets. The results show that the proposed double-threshold scheme can effectively improve accuracies of the converted SNNs. More importantly, the proposed AugMapping is more advantageous for constructing accurate, fast and efficient deep SNNs as compared to other state-of-the-art approaches. Our study therefore provides new approaches for further integration of advanced techniques in ANNs to improve the performance of SNNs, which could be of great merit to applied developments with spike-based neuromorphic computing.


\end{abstract}

%
\IEEEpeerreviewmaketitle

\section{Introduction}

\IEEEPARstart{A}{s} a subset of artificial neural networks (ANNs), deep neural networks (DNNs) \cite{lecun2015deep} have shown significant improvements in a wide range of tasks such as image classification \cite{krizhevsky2012imagenet}, speech recognition \cite{xiong2017toward}, natural language processing \cite{hirschberg2015advances} and robotics \cite{silver2017mastering}, etc. However, with the complexity of neural networks increasing progressively, running such deep networks often requires large amounts of computational resources such as memory and power, thus limiting their applied developments in battery-constrained devices such as cell phones and embedded electronics \cite{severa2019training}. Some studies focus on reducing the network connections and using low-precision parameters \cite{han2015learning, courbariaux2015binaryconnect,rastegari2016xnor,zhu2017trained}, but the computational consumption is still large.  Hence, the challenge still remains for low-power paradigms to enlarge the applicability of DNNs.

Different from numerical values used by traditional ANNs, spiking neural networks (SNNs) emulates the brain in a way to utilize discrete spikes for information representation and transmission, and thus are more brain-like and computationally powerful \cite{maass1997networks,kandel2000principles}. Moreover, owing to the discrete feature of spikes over an additional time dimension, SNNs are capable of asynchronous and sparse computation under an event-based manner where a computational budget is paid only at the appearance of a spike event \cite{nawrocki2016a,yu2018spike, roy2019towards,feldmann2019all}. This motivates the development of neuromorphic computing platforms that have successfully shown a remarkable performance of orders of magnitude more efficient in terms of power consumption than conventional computing platforms \cite{furber2014spinnaker, merolla2014million, davies2018loihi}. The attractive potential of spike-based computation is drawing increasing efforts to the development of SNNs \cite{gutig2006tempotron, yu2013rapid, yu2015spiking, taherkhani2018supervised,hong2019training}, which yet is still very much in its infancy.

Although SNNs are promising for low-power and fast inference by their nature, training such deep networks is difficult due to their inherent characteristics of discontinuity, non-linearity and complexity on dynamics, let alone to achieve a competitive recognition accuracy to ANNs \cite{pfeiffer2018deep,tavanaei2019deep,taherkhani2020a}. There are two mainstream approaches developed to overcome the challenge of training deep SNNs: direct and indirect training. The indirect training approaches are also often referred as conversion or mapping methods.

An early attempt of the direct training methods is based on spike-timing-dependent-plasticity (STDP) where synaptic modification is controlled by the local correlation of pre- and postsynaptic spike timings \cite{bi1998synaptic}. However, SNNs trained with STDP are normally limited to shallow structures, and cannot be scaled up to large networks with high performances due to the lack of a global instructor \cite{diehl2015unsupervised,mozafari2018first, kheradpisheh2018stdp,lee2018deep}. Backpropagation (BP), a prevalent learning algorithm in ANNs \cite{rumelhart1988learning}, has been successfully applied to train deep SNNs by addressing the inherent non-differentiable obstruction during the backward propagation of an error instructor \cite{bohte2002error, mostafa2017supervised,wu2018spatio, jin2018hybrid, shrestha2018slayer, neftci2019surrogate, gu2019stca}. 
A surrogate is usually designed to approximate the gradients in these BP-based approaches. This approximation is effective with relatively shallow structures,
but normally gets worse for more challenging tasks and deeper networks.
Additionally, direct training methods are often time-consuming during the adaptation of neural parameters.
Moreover, as compared to ANNs, there is still a big gap with respect to the recognition accuracy for these direct training methods \cite{pfeiffer2018deep,tavanaei2019deep}.

\begin{figure}
	\centering\includegraphics[width=0.47\textwidth]{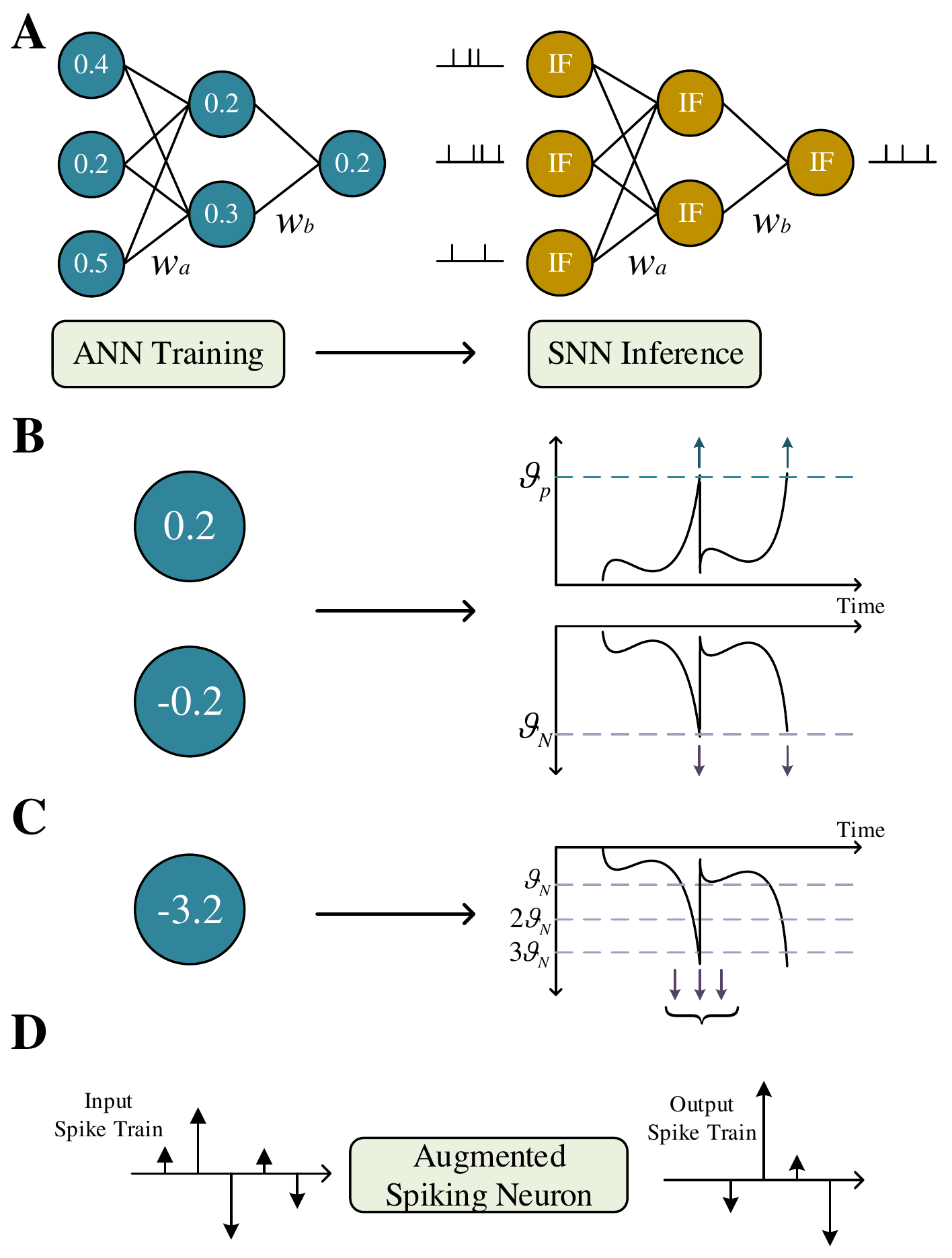}
	\caption{Illustrations of ANN-to-SNN conversion. \textbf{A}, a standard conversion scheme. \textbf{B}, mapping both positive and negative activations in ANNs with a double-threshold scheme. \textbf{C}, augmented spikes to address the over-activation issue. \textbf{D}, augmented spiking neuron that receives and elicits augmented spikes.}
	\label{illustration}
\end{figure}

In order to narrow the gap between SNNs and ANNs, a conversion scheme emerges by mapping the weights of a trained ANN to an SNN of the same architecture (see Fig.~\ref{illustration}\textbf{A}). This conversion scheme leverages on advanced techniques in ANNs to achieve a comparable accuracy with SNNs. 
One of the most early works successfully introduces the conversion scheme with complicated spiking neurons, but suffers from a significant accuracy loss \cite{perez2013mapping}. Later efforts discover that the firing rates of spiking neurons can approximate the activations of their counterparts in ANNs with sufficient time steps \cite{cao2015spiking}. This finding has become the fundamental principle underlying the conversion scheme. A shallow convolutional network trained with certain constraints can be successfully deployed to an SNN, resulting in a good accuracy on traditional object recognition benchmarks \cite{cao2015spiking}. Then, a data-based normalization (DataNorm) \cite{diehl2015fast} is developed to achieve a nearly lossless performance by threshold-balancing for a proper information transmission with firing rates, but the techniques used for the conversion are rather limited. An extended variant \cite{rueck2017conversion} is thus developed later for further improvements by incorporating more techniques from ANNs, including biases, max pooling, softmax and batch normalization.
Later, SpikeNorm \cite{sengupta2019going} is developed to scale neural parameters according to the activities of SNNs rather than ANNs, leading to a nearly lossless conversion even in very deep networks.
Recently, a channel-wise normalization is introduced to further minimize the conversion loss with an elaborate adjustment, and achieves comparable results to ANNs on the object detection task \cite{kim2020spiking}.
 
Despite remarkable achievements of nearly lossless conversion even in very deep networks, threshold-balancing techniques \cite{diehl2015fast,rueck2017conversion,sengupta2019going, kim2020spiking} require a large set of training data to extract auxiliary values used in the conversion. Therefore, these techniques are data-driven and susceptible to the selection of samples. Furthermore, the current conversion approaches encounter other common challenges as follows:

\begin{itemize}
	\item A delicate balance between weights and thresholds to avoid information loss caused by over- or under-activation \cite{diehl2015fast} makes the current conversion approaches quite complicated. 
	
	\item The converted SNNs with current approaches are inefficient in both time and energy as they require a large number of spikes and time steps especially for significantly deep networks.
	
	\item Most of approaches cannot convert negative activations in ANNs to their spiking counterparts, and thus advanced variations \cite{xu2015empirical} like LeakyReLU are unable to be utilized. This limits the potential of SNNs to take full advantages of advanced techniques in ANNs.
\end{itemize}

In order to improve the inference speed and energy efficiency, preliminary efforts are made to extend spikes with different forms, such as multi-strength spikes \cite{chen2018fast}, weighted spikes \cite{kim2018deep} and burst spikes \cite{park2019fast}. Despite their efficacy in reduction of classification latency and number of events, these works are still limited to constrained ANNs with positive activations only.
Moreover, their conversions are based on either complicated schemes like DataNorm or an exhaustive strategy with manual exploration. A clear and simple rule to determine proper parameters for spiking neurons is still under exploring.

In this paper, we first propose a conversion approach called TerMapping by extending DataNorm \cite{diehl2015fast} with a new double-threshold scheme. 
Then, we introduce an advanced neuron model that is capable of receiving and eliciting augmented spikes. Based on the model, another new conversion method called AugMapping is developed.
We evaluate the performance of our methods with extensive experiments. Our major contributions can be enumerated as follows:

\begin{itemize}
	\item A double-threshold scheme is introduced such that SNNs are enhanced to represent both positive and negative activations, relieving constraints on the selection of ANNs.	
	
	\item A new scheme of augmented spike is introduced where a spike coefficient is used to represent additional information including both the polarity and the number of binary spikes occurring at one time step. This  enables the correspondingly developed augmented neurons to completely overcome the pathological phenomenon of over-activation \cite{diehl2015fast} on one hand, and to improve both the accuracy and efficiency with less number of spike events on the other hand.
		
	\item New conversion methods are developed with a clear and simple rule to assign suitable parameters for SNNs. More importantly, a detailed theoretical formulation about the approximation is provided, supporting the efficacy of our methods and paving the way for other related future works.
	
	\item We evaluate the effectiveness of our methods with various network structures based on different datasets including MNIST, FashionMNIST and CIFAR10. Experimental results show that our methods can achieve nearly lossless ANN-to-SNN conversion, and importantly are more fast, accurate and efficient than the current state-of-the-arts.	
	Our work thus contributes to improve the performance of spike-based inference, which would be of great merit to neuromorphic computing.
\end{itemize}

The remainder of this paper is organized as follows. Section~\ref{secMet} presents the details of our proposed methods, followed by the experimental results and discussions in Section~\ref{secExp} and Section~\ref{SecDis}, respectively. Finally, a conclusion is provided in Section~\ref{secCon}.

\section{Methods}
\label{secMet}
\subsection{Double-Threshold Spiking Neural Networks}
Direct training of SNNs to achieve competitive accuracies with ANNs is still challenging. Hence, a straightforward approach emerges by converting the weights of a pre-trained ANN to their counterparts in an SNN of the same structure. 
However, the setup of a single threshold in standard spiking neurons makes them only capable of representing information of a sole polarization. 
This makes it difficult to map negative activations from ANNs to SNNs, and thus constraints are often applied to ANNs to keep only positive activations. Such a constraint can decrease the performance of the pre-trained ANN, and so as the converted SNN.

In order to address this limitation, we introduce a double-threshold firing scheme (see Fig.~\ref{illustration}\textbf{B}) with the integrate-and-fire (IF) model \cite{gerstner2002spiking}, one of the most widely studied neuron models in various conversion approaches \cite{cao2015spiking,diehl2015fast,rueck2017conversion,sengupta2019going}.  
Each spiking neuron maintains an internal state called membrane potential and continuously integrates postsynaptic potentials generated by afferent spikes into its membrane potential. The integration dynamics of the $i$-th neuron in the $l$-th layer at time $t$ is described as:
\begin{align}
V_i^l(t) = V_i^l(t-1) + \sum_{j=1}^{M^{l-1}}w_{ij}^lo_j^{l-1}(t) + b_i^l
\label{eq1}
\end{align}%
where ${V_i^l(t)}$ represents the membrane potential, and ${M^{l-1}}$ denotes the number of neurons in the preceding layer. ${o_j^{l-1}(t)}$ is the afferent spike, and ${w_{ij}^l}$ is the weight connection from the $j$-th neuron in layer $l-1$ to the $i$-th one in layer $l$. ${b_i^l}$ is the corresponding bias.

Once the membrane potential of the neuron crosses either the positive threshold ${\vartheta_P}$ or the negative one ${\vartheta_N}$ (see Fig.~\ref{illustration}\textbf{B}), a corresponding polarized spike will be elicited and propagated to downstream neurons, as formulated by:

\begin{align}
o_{i}^{l}(t)=
\begin{cases}1 &  if\ \ V_{i}^{l}(t)\geq{\vartheta_P}\\
-1 & if\ \ V_{i}^{l}(t)\leq{\vartheta_N}\\
0 & otherwise
\end{cases}
\label{eq2}
\end{align}%

At the occurrence of a polarized spike, the neuron instantaneously triggers a reset process where its membrane potential is changed by a value of the corresponding threshold, as given by: 
\begin{align}
V_{i}^{l}(t)=\begin{cases}
V_{i}^{l}(t)-{\vartheta_P} & if\ \ o_{i}^{l}(t)= 1  \\
V_{i}^{l}(t)-{\vartheta_N} & if\ \ o_{i}^{l}(t)= -1  \\
V_{i}^{l}(t) &otherwise
\end{cases}
\label{eq3}
\end{align}

Based on the above neuron model, a double-threshold SNN can thus be easily constructed.

\subsection{TerMapping}
A fundamental guideline for the conversion approaches is that the firing rates of spiking neurons need to approximate the activations of their counterparts in an ANN. 
A standard spike form of all-or-nothing could inevitably lead to information loss during conversion due to its limited capacity for transmission at each time step. Both over-activation and under-activation in ANNs can result in improper representation with spikes, thus decreasing recognition performance of SNNs \cite{diehl2015fast}.
Appropriate balance between thresholds and input firing rates in SNNs provides an effective approach to relieve the loss to a certain extent, and can even achieve nearly lossless conversion with an elaborate configuration \cite{cao2015spiking,diehl2015fast,rueck2017conversion,sengupta2019going}.
However, most of the current conversion approaches rely on the requirements of only positive activations, limiting the selection of ANNs to be probably sub-optimal. Our double-threshold scheme provides a solution to address this issue by incorporating an additional negative threshold to spiking neurons. Our double-threshold scheme can be applied to different conversion approaches, and here we select DataNorm \cite{diehl2015fast} as a representative for extension, based on which we develop a new method named TerMapping.

\newcommand\mycommfont[1]{\small\ttfamily\textcolor{black}{#1}}
\SetCommentSty{mycommfont}

\begin{algorithm}
	\caption{Computation of Scaling Factors in TerMapping} 
	\label{alg:BinMapping} 
	\SetKwFunction{isOddNumber}{isOddNumber}
	\SetKwInOut{KwIn}{Input}
	\SetKwInOut{KwOut}{Output}
	
	\KwIn{The number of layers $n$, each layer's weights $w_{i}$ and output activations $z_{i}$, $i=1,2,...,n$.}
	\KwOut{Corresponding scaling factors $\lambda_i$.}
	
	$pre\_factor = 1$\\
	\For{$i \leftarrow 1$ \KwTo $n$}{
		\tcc{Compute the maximum absolute value of weights and output activations}
		$max\_weight = w_i.abs().max()$\\
		$max\_output = z_i.abs().max()$\\
		$post\_factor = max(max\_weight, max\_output)$\\
		\tcc{Obtain the scaling factor for the current layer }
		$\lambda_i = post\_factor / pre\_factor$\\
		$pre\_factor = post\_factor$\\
	}
\end{algorithm}

After an ANN has been trained, the training set is fed to it again, and the maximum absolute values of both output activations and weights in every layer are recorded to obtain the scaling factors used for balancing procedures. In the inference with the converted SNN, firing thresholds in each layer are rescaled by their corresponding scaling factors recorded from the previous step. 
Our method is effective to control the firing rates of most converted neurons in a normalized range between 0 and 1, thus reducing the accuracy loss caused by improper activations. Pseudo-codes for the computation of scaling factors are shown in Algorithm~\ref{alg:BinMapping}.

\subsection{Augmented Spikes}

Due to the representation constraint with all-or-nothing spikes at each time step, a delicate balance between thresholds and firing rates is required to reduce information loss. For example, if over-activation happens to result in more than one spike in a single time step, the standard spiking neuron can only elicit maximally one accordingly, thus inevitably leading to the decrease in performance. In order to address this issue, we introduce a new scheme of augmented spike where a spike coefficient is employed to carry additional information including both polarity and the number of typical spikes occurring at one time step (see Fig.~\ref{illustration}\textbf{C}). 
Specifically, the presenting form of $o_{i}^{l}(t)$ at a time step is extended from binary to multiple stages.
Augmented spikes extend the capability of spike-based representation, and thus could be useful to reduce information loss in conversion approaches.

Endowing spiking neurons with the ability of processing and eliciting augmented spikes, a new augmented spiking neuron model is developed (see Fig.~\ref{illustration}\textbf{D}). Whenever a firing condition is reached, the neuron will elicit an augmented spike, as:
\begin{align}
o_{i}^{l}(t)=\left\{ \begin{array}{*{35}{l}}
\left\lfloor \frac{V_{i}^{l}(t)}{{\vartheta_P}} \right\rfloor  & if\ \ V_{i}^{l}(t)\geq{\vartheta_P}  \\
-\left\lfloor \frac{V_{i}^{l}(t)}{{\vartheta_N}} \right\rfloor  & if\ \ V_{i}^{l}(t)\leq{\vartheta_N}  \\
0 & otherwise  \\
\end{array} \right.
\label{eq4}
\end{align}%
where the floor division operator ${\lfloor . \rfloor}$ returns the integer value of the quotient. 

If an augmented spike is elicited, the membrane potential of the spiking neuron is instantaneously decreased or increased by a certain amount of the corresponding threshold levels, as given by:
\begin{align}
V_{i}^{l}(t)=\begin{cases}
V_{i}^{l}(t)-o_{i}^{l}(t){\vartheta_P} & if\ \ o_{i}^{l}(t)\geq 1  \\
V_{i}^{l}(t)+o_{i}^{l}(t){\vartheta_N} & if\ \ o_{i}^{l}(t)\leq -1  \\
V_{i}^{l}(t) & otherwise \\
\end{cases}
\label{eq5}
\end{align}

\subsection{AugMapping}
\label{secAug}
Based on the augmented spiking neuron model, we propose a new AugMapping method to realize ANN-to-SNN conversion. 
The augmented firing scheme enables spiking neurons to represent both positive and negative activations of an ANN, being beneficial to adopt more advanced activation functions like LeakyReLU \cite{xu2015empirical, maas2013rectifier} for a better recognition accuracy with the converted SNN.
More importantly, complicated balancing techniques required for a proper transmission of information with standard spikes can be eliminated under our augmented scheme thanks to its advanced capacity for information representation.
As a result, a more simple and clear technique can be developed for direct construction of an SNN with a pre-trained ANN, as detailed in the following.

In our AugMapping, the firing rate of an augmented spiking neuron should approximate the output activation of its counterpart in ANN. Here, we present an analytical description for this approximation, and on its basis, we can derive a simple rule to assign appropriate parameters for the converted spiking neurons.

The output activation of a neuron in ANNs with LeakyReLU or ReLU activation function can be formulated by:
\begin{align}
z_{i}^{l}=\alpha (\sum\limits_{j=1}^{{{M}_{l-1}}}{w_{ij}^{l}}z_{j}^{l-1} + b_i^l)
\label{eq6}
\end{align}%
where ${z_i^l}$ indicates the output activation of the $i$-th neuron in the $l$-th layer. ${z_j^{l-1}}$ and ${b_i^l}$ are the corresponding input and bias, respectively. ${\alpha}$ is a coefficient controlling the slope of the activation function. For simplicity, we fix the bias to zero in both ANN and SNN, as is similar to other previous works \cite{cao2015spiking,diehl2015fast}. 

We analyze the correlation between the output activations in the ANN and the firing status in the SNN. Each input pattern is presented for a total number of $T$ time steps. Note that, we only describe the case where the neuron crosses its positive threshold for the sake of simplicity, while a similar procedure can be easily applied to describe the negative one. 

From an initial position of zero, the membrane potential at ${T}$ can be obtained by recursively applying Eq.~(\ref{eq1}) and Eq.~(\ref{eq5}), given by:
\begin{align}
	V_{i}^{l}(T)=\sum\limits_{j=1}^{{{M}^{l-1}}}{w_{ij}^{l}}\sum\limits_{{{t}^{{}}}=1}^{T}{o_{j}^{l-1}}({{t}^{{}}})-{\vartheta_{P}}_{{}}\sum\limits_{{{t=1}^{{}}}}^{T-1}{o_{i}^{l}}(t)
	\label{eq8}
\end{align}%

According to Eq.~(\ref{eq4}), this potential can also be given as:
\begin{align}
V_{i}^{l}(T)={\vartheta_{P}}(o_{i}^{l}(T)+\sigma)
\label{eq9}
\end{align}%
where ${\sigma}$ is a residual item, given by ${\sigma =\frac{V_{i}^{l}(T)}{{\vartheta_{P}}}-\left\lfloor \frac{V_{i}^{l}(T)}{{\vartheta_{P}}} \right\rfloor}$. 

Taking Eq.~(\ref{eq8}) and Eq.~(\ref{eq9}), we can futher get:
\begin{align}
{\vartheta_{P}}(\sum\limits_{t=1}^{T}{o_{i}^{l}}(t)+\sigma) =\sum\limits_{j=1}^{{{M}^{l-1}}}{w_{ij}^{l}}\sum\limits_{t=1}^{T}{o_{j}^{l-1}}(t)
\label{eq10}
\end{align}%

In order to correlate the output activation of the neuron in ANN with the firing rate of the one in SNN, we define the firing rate as 
\begin{align}
{r_{i}^{l}(T)=\frac{N_{i}^{l}(T)}{T}=\frac{\sum\limits_{t=1}^{T}{o_{i}^{l}}(t)}{T}}
\label{eqrate}
\end{align}
where ${N_i^l(T)}$ is the number of spikes generated during the total $T$ time steps. Eq.~(\ref{eq10}) can thereafter be converted into:
\begin{align}
{\vartheta_{P}}r_i^l(T) =\sum\limits_{j=1}^{{{M}^{l-1}}}{w_{ij}^{l}}r_j^{l-1}(T) - \frac{\sigma\vartheta_{P}}{T}
\label{eq11}
\end{align}%

Eq.~(\ref{eq11}) is also a recursive expression, based on which we can take approximations layer by layer. In the first hidden layer, for simplicity, the inputs of both ANN and SNN are identical, which satisfies ${r^0(T)=z^0}$. Equating Eq.~(\ref{eq6}) and Eq.~(\ref{eq11}) yields:%
\begin{align}
r_{i}^{1}(T)=\frac{\text{z}_{i}^{1}}{\alpha {\vartheta_{P}}}-\frac{\sigma }{T}
\label{eq12}
\end{align}%
which indicates that as ${T\rightarrow \infty}$, the firing rate ${r_i^1(T)}$ approaches to its target value ${z_i^1}$ when ${\alpha\vartheta_{P} = 1}$.

Augmented neurons in higher layers continuously integrate spikes from their preceding layer. We can evaluate the approximation errors in higher layers with the recursive expression of Eq.~(\ref{eq11}). Incorporating the first layer as described by Eq.~(\ref{eq12}), the firing rates in higher layers can be given as:%
\begin{align}
r_{i}^{l}(T)=\frac{z_{i}^{l}}{{{(\alpha  {\vartheta_{P}})}^{l}}}-\frac{\sigma }{T }(1+\sum\limits_{n=2}^{l}{\prod\limits_{{{l}^{\prime }}=n}^{l}{\frac{\sum\limits_{j=1}^{{{M}^{{{l}^{\prime }}-1}}}{w_{ij}^{{{l}^{\prime }}}}}{{\vartheta_{P}}}}})
\label{eq13}
\end{align}%

Eq.~(\ref{eq13}) implies that the approximation error accumulates in deeper layers, and as a result, more time steps are required for lossless conversion as the network structure becomes deeper.

Both Eq.~(\ref{eq12}) and Eq.~(\ref{eq13}) suggest a correlation between the firing threshold and the activation slope, i.e. ${\vartheta_P = \frac{1}{\alpha}}$, to make the firing rate of the neuron in SNN approach to the output activation of that in ANN. 
Similarly, the negative firing threshold can be determined by ${\vartheta_N = -\frac{1}{\alpha}}$. Therefore, the firing thresholds of converted spiking neurons can be assigned with a clear and simple rule according to our analytical descriptions rather than exhaustive manual explorations or time-consuming configurations.

Similar to the routines of a typical conversion scheme, our AugMapping will convert a pre-trained ANN to an SNN, but differently with a more direct and simple method. Importantly, our theoretical analysis supports that the as-proposed AugMapping is able to achieve nearly lossless conversion under certain conditions.

\section{Experimental Results}
\label{secExp}

\subsection{Experimental Setup}

The performance of our methods is extensively examined with six different networks based on various datasets including MNIST \cite{lecun1998gradient}, and the more challenging Fashion-MNIST \cite{xiao2017/online} and CIFAR10 \cite{krizhevsky2009learning}.

MNIST is a handwritten digit image dataset and consists of 60,000 images for training and 10,000 images as the test set. Each sample labeled 0-9 is a grayscale image with a size of
28${\times}$28. Differently, Fashion-MNIST is a rather new dataset with different classes of clothing. It shares the same image size and structure of training and testing splits with MNIST, but is a more challenging image classification task. CIFAR10 contains 60,000 color images belonging to 10 classes. Each image consists of 32${\times}$32 pixels. No data augmentation is applied on MNIST and Fashion-MNIST. For CIFAR10, we utilize a standard augmentation where each training sample are padded with 4 pixels on each side of the image, followed by a 32${\times}$32 crop and a random horizontal flip. 

\begin{table}[!htb]
	\centering
	\caption{Experimental network configurations}
	\setlength{\tabcolsep}{3mm}{
		\begin{tabular}{Sc||Sc Sc}
			\hline
			Dataset       & Network  & Topology                \\ \hline
			\multirow{2}{*}{MNIST}   & Net1 & 1200-1200-10            \\
			& Net2 & 12c5-p2-64c5-p2-10      \\ \hline
			\multirow{2}{*}{Fashion-MNIST} & Net3 & 6400-10                 \\
			& Net4 & 32c5-p2-64c5-p2-1024-10 \\ \hline
			\multirow{2}{*}{CIFAR10} & Net5 & \begin{tabular}[c]{@{}c@{}}128c3-128c3-p2-256c3-256c3\\ -p2-512c3-512c3-p2-1024-10\end{tabular} \\
			& Net6 & VGG16 \cite{simonyan2014very}                   \\ \hline
		\end{tabular}
	}
	\label{table1}
\end{table}

As is shown in Table~\ref{table1}, both types of fully connected (Net1 and Net3) and convolutional networks (Net2 and Net4) are applied for MNIST and Fashion-MNIST. Two deeper networks with the VGG architecture (Net5 and Net 6) are used for CIFAR10. 
The detailed network structures are shown in the table, where c and p represent the convolutional and pooling layer, respectively. 
For example, 15c5 stands for a convolutional layer with 15 feature maps of 5${\times}$5 kernel size, and p2 denotes a pooling layer with a receptive window of 2${\times}$2. A single digit such as 250 represents a fully connected layer with 250 neurons. 
In VGG networks, a dropout layer is used after every LeakyReLU layer except for those layers which are followed by a pooling one.

\begin{figure*}[!htb]
	\centering\includegraphics[width=0.94\textwidth]{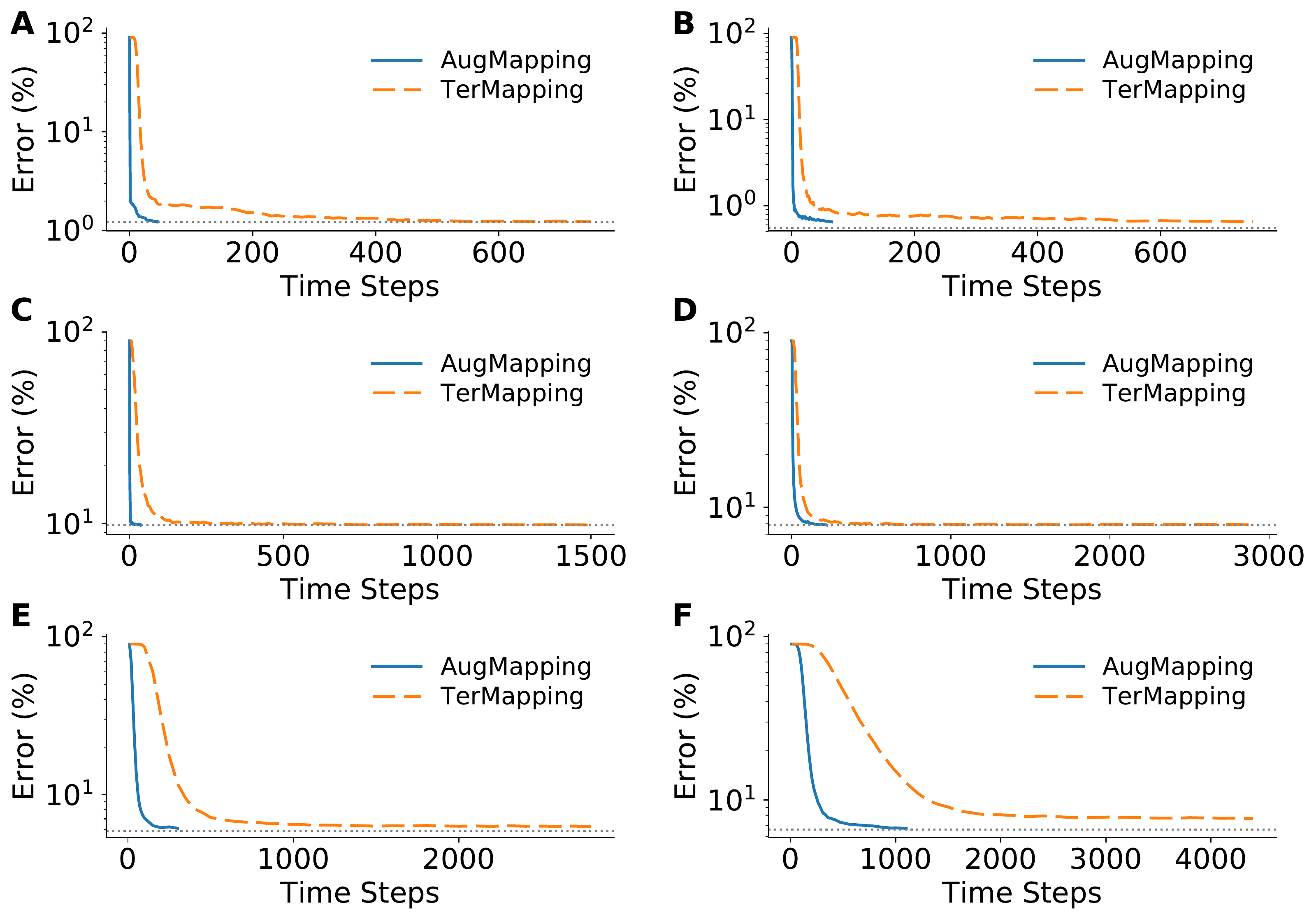}
	\caption{Inference error versus time steps. Results for Net1 to Net6 are presented in pannel \textbf{A} to \textbf{F} accordingly. The horizontal dotted lines in each panel denote the accuracies of the corresponding ANNs.}
	\label{cmp_speed}
\end{figure*}

Both the training of ANNs and the inference with converted SNNs are implemented with the mainstream framework PyTorch \cite{paszke2019pytorch}.
During inference with SNNs, pixel values of images are directly fed into the first hidden layer in order to remove variability \cite{rueck2017conversion}. The categorical decision made by the output layer is determined by the unit that has the biggest firing rate.

\subsection{Results}

In order to test the efficacy of our double-threshold scheme, we first compare our TerMapping with DataNorm \cite{diehl2015fast} under the same conditions. 
As is shown in Table~\ref{table2}, our TerMapping is more accurate than DataNorm thanks to its extended ability with the double-threshold scheme to represent both positive and negative activations in advanced ANNs. 

\begin{table}[!htb]
	\centering
	\caption{Accuracy compasiron between DataNorm and TerMapping}
	\setlength{\tabcolsep}{4mm}{
		\begin{tabular}{Sc||Sc Sc Sc}
			\hline
			Network  & Method     & ANN acc. & SNN acc. \\ \hline
			\multirow{2}{*}{Net1} & DataNorm \cite{diehl2015fast}  & 98.68\%  & 98.64\%  \\
			& \textbf{TerMapping} & 98.77\%  & \textbf{98.77\%}  \\ \hline
			\multirow{2}{*}{Net2} & DataNorm \cite{diehl2015fast}  & 99.14\%  & 99.10\%  \\
			& \textbf{TerMapping} & 99.35\%  & \textbf{99.35\%}  \\ \hline
		\end{tabular}
		\label{table2}
	}
\end{table}

Then, we focus more on investigating the efficacy of our augmented scheme by providing more detailed comparisons between TerMapping and AugMapping. Different measurement metrics are presented in Table~\ref{table3}.

\begin{table}[!htb]
	\centering
	\caption{Conversion results with our methods of TerMapping and AugMapping}
	\begin{tabular}{Sc Sc||Sc Sc Sc Sc}
		\hline
		Network &
		\begin{tabular}[c]{@{}c@{}}ANN \\ acc.\end{tabular} &
		Method &
		\begin{tabular}[c]{@{}c@{}}SNN \\ acc.\end{tabular} &
		Latency &
		\begin{tabular}[c]{@{}c@{}}\# Events\\ ($10^6$)\end{tabular} \\ \hline
		\multirow{2}{*}{Net1} & \multirow{2}{*}{98.77\%} & TerMapping & 98.77\% & 750  & 0.03  \\
		&                          & AugMapping & 98.77\% & 46   & 0.02  \\ \hline
		\multirow{2}{*}{Net2} & \multirow{2}{*}{99.35\%} & TerMapping & 99.35\%   & 750  & 0.28  \\
		&                          & AugMapping & 99.35\%   & 65   & 0.11  \\ \hline
		\multirow{2}{*}{Net3} & \multirow{2}{*}{90.18\%} & TerMapping & 90.18\% & 1500 & 0.05  \\
		&                          & AugMapping & 90.18\% & 37   & 0.03  \\ \hline
		\multirow{2}{*}{Net4} & \multirow{2}{*}{92.11\%} & TerMapping & 92.11\% & 2900 & 2.45  \\
		&                          & AugMapping & 92.11\% & 220  & 0.68  \\ \hline
		\multirow{2}{*}{Net5} & \multirow{2}{*}{94.13\%}   & TerMapping & 93.75\% & 2800 & 15.36 \\
		&                          & AugMapping & 93.90\%  & 300  & 12.51 \\ \hline
		\multirow{2}{*}{Net6} & \multirow{2}{*}{93.42\%}   & TerMapping & 92.30\%    & 4400 & 17.66 \\
		&                          & AugMapping & 93.29\%   & 1100 & 18.10 \\ \hline
	\end{tabular}
	\label{table3}
\end{table}

As can be seen from the table, both AugMapping and TerMapping can achieve a nearly lossless conversion for various types of networks ranging from shallow to deep structures over different datasets. For the relatively simple task of MNIST where shallow networks are sufficient enough, both of our two methods can achieve the same test accuracies as the corresponding ANNs, while the AugMapping has much shorter classification latency and consumes significantly smaller number of events than the TerMapping, reflecting the efficiency in both time and energy. On the more challenging Fashion-MNIST dataset, both our methods still succeed in no-accuracy-loss conversions with Net3 and Net4. 
Note that our methods significantly outperforms the recent reported result such as in \cite{hao2020biologically} (around 85\%), highlighting our contribution in improving the spike-based performance in accuracy. 

As the task gets more challenging in CIFAR10 where significantly deep networks are adopted, reaching a lossless accuracy is getting more difficult for both of our methods within thousands of time steps. Nevertheless, our AugMapping consistently outperforms the other one with a better accuracy and lower latency.
Notably, with the networks getting deeper, more events and time steps are required to achieve a nearly lossless conversion with both of our methods, being consistent with our theoretical analysis provided in Section~\ref{secAug}.

\begin{figure*}[!htb]
	\centering\includegraphics[width=0.94\textwidth]{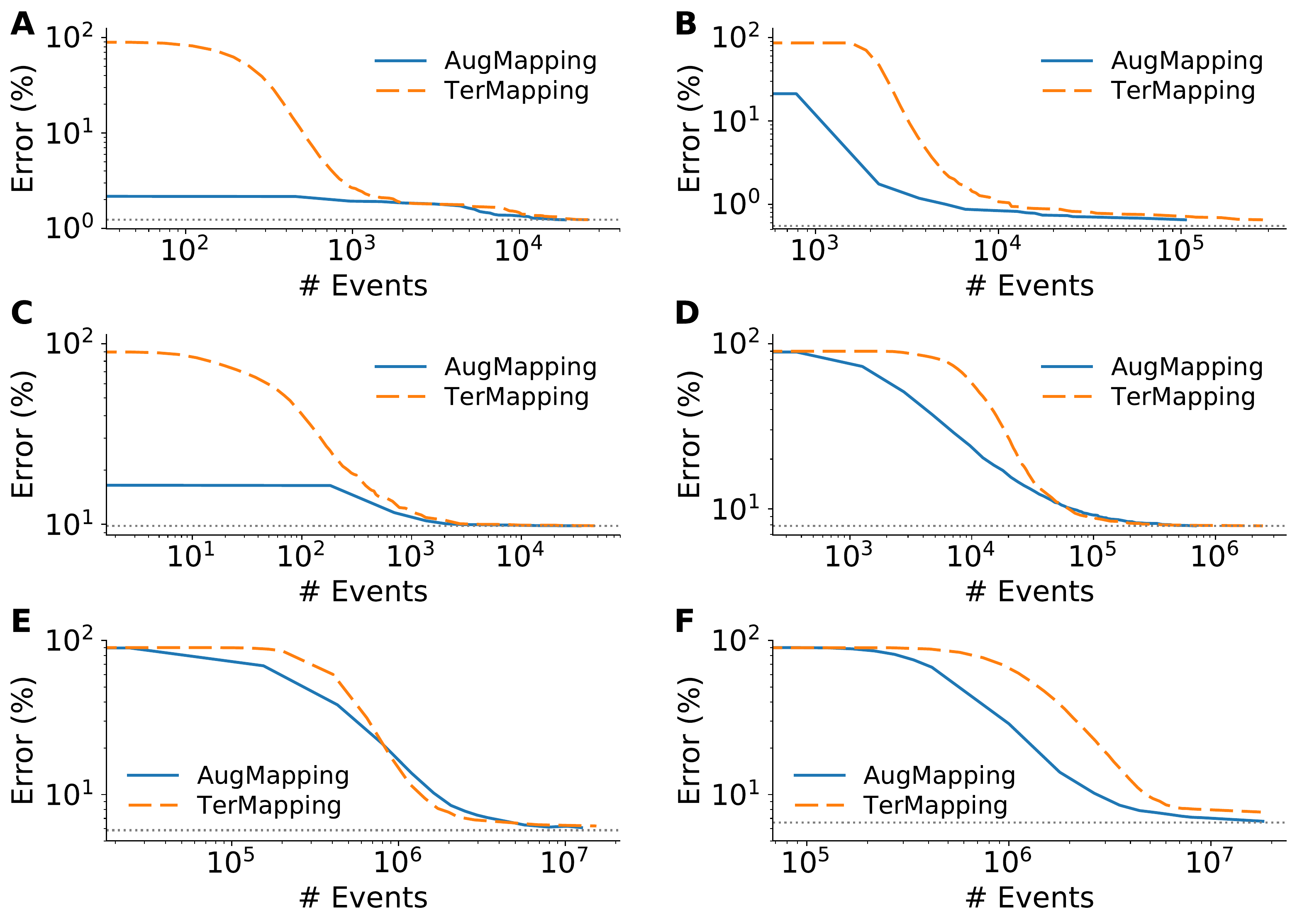}
	\caption{Inference error versus the number of events in SNNs. \textbf{A} to \textbf{F} shows the results of Net1 to Net6 accordingly. The horizontal dotted lines in each panel represent the accuracies of the corresponding ANNs.}
	\label{cmp_events}
\end{figure*}

The inference errors of converted SNNs decrease as the number of time steps increases. The evolving details for both AugMapping and TerMapping are shown in Fig.~\ref{cmp_speed}. 
As can be seen, the AugMapping is significantly more faster than the TerMapping to reach a nearly lossless accuracy for all the networks.
In most cases, the AugMapping is faster than the other one with at least one order of magnitude, while achieves a better accuracy.
As the firing ability of the neuron at each time step is rather constrained in TerMapping to relieve issues of over-activation and under-activation, spiking neurons normally need a large number of time steps to precisely reflect the activation of their counterparts in ANNs. When networks get deeper, significantly more time steps are required for SNNs to accumulate sufficient information through layers. In the contrary, the AugMapping only takes a small number of time steps to achieve a remarkable accuracy, owing to the enhanced capability for information representation with augmented spikes. As a result, the AugMapping is advantageous in accuracy, speed and energy-consumption.

Additionally, we perform a detailed examination on the total number of spike events consumed to achieve certain levels of accuracies. As can be seen from 
Fig.~\ref{cmp_events}, there is a trade-off between the accuracy and the number of events for both methods across all networks.
To be more specific, a high accuracy requires a large number of events, indicating a sacrifice on energy efficiency.
Gladly, the headache on energy consumption can be relieved by setting an acceptable level for accuracies. Notably, the AugMapping often generates a smaller number of events than that of the TerMapping in most cases, indicating its high energy-efficiency since the power consumption of SNNs is roughly proportional to the number of events \cite{cao2015spiking,kim2018deep,krithivasan2019dynamic}. 

The above results highlight the effectiveness of both our double-threshold and augmented schemes, which play an important role in improving the performances of SNNs.


\subsection{Early Decision}

As is compared to an ANN, SNN is advantageous in a prompt response thanks to a fast and asynchronous propagation of spikes through the network \cite{pfeiffer2018deep}, and as a result, decisions can be made based on early spikes.
In this part, we examine the property of our methods with respect to early decisions at the cost of a certainly small loss on accuracy. Notably, a small sacrifice on accuracy will offer the opportunity for a great level of speed acceleration and spike event reduction.
In our experiments, different levels of accuracy loss are used to closely evaluate the benefits on the reduction of both latency and spike events.

\begin{figure}[!htb]
	\centering\includegraphics[width=0.47\textwidth]{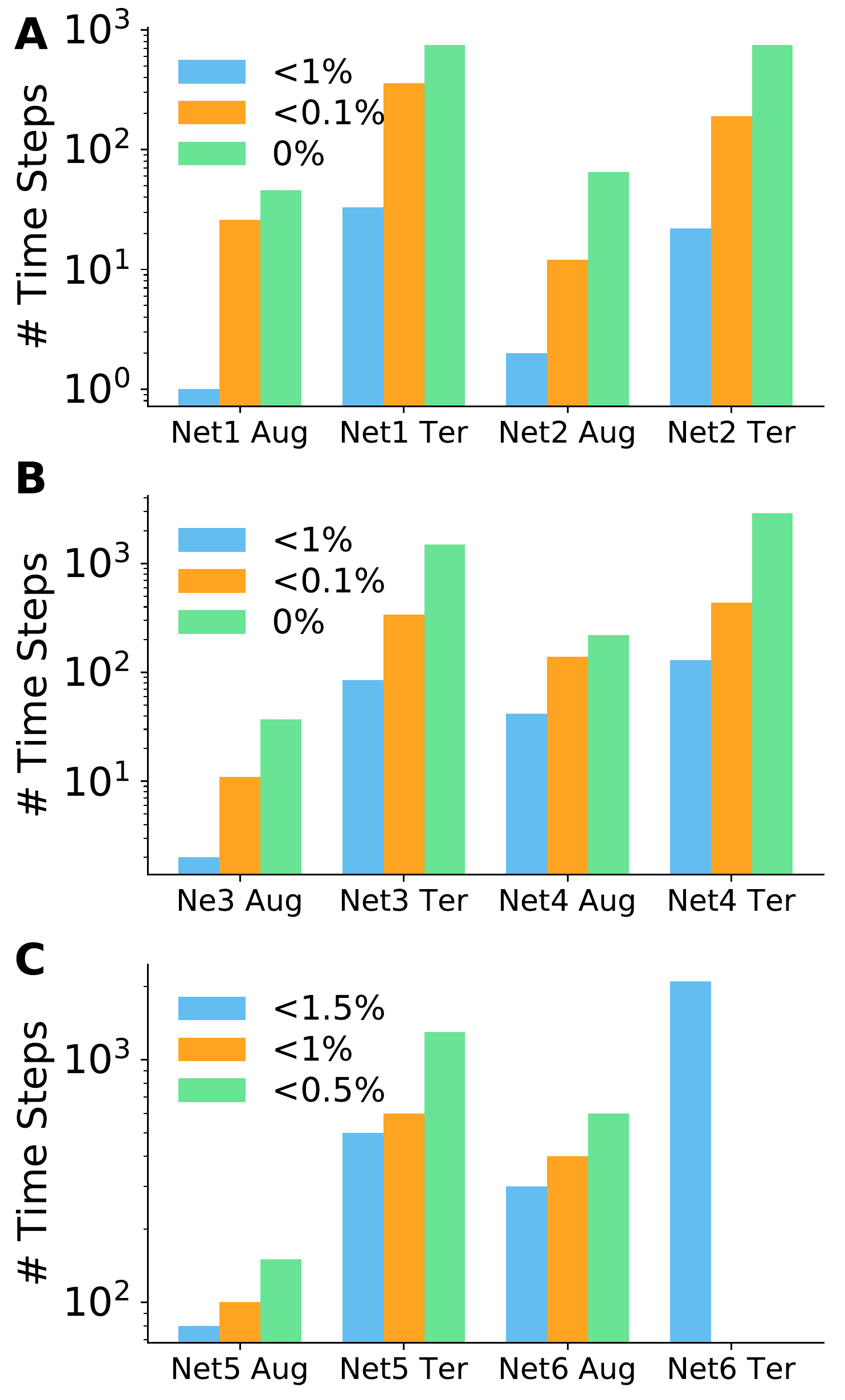}
	\caption{The number of required time steps to reach an acceptable accuracy specified by certain loss conditions for early decisions. For example, a condition like `$<1\%$' represents the maximally allowed accuracy loss from the ANN baseline is 1\%. Some results for TerMapping on Net6 are not shown as no convergence appears within thousands of time steps.}
	\label{early_decision_latency}
\end{figure}

\begin{figure}[!htb]
	\centering\includegraphics[width=0.47\textwidth]{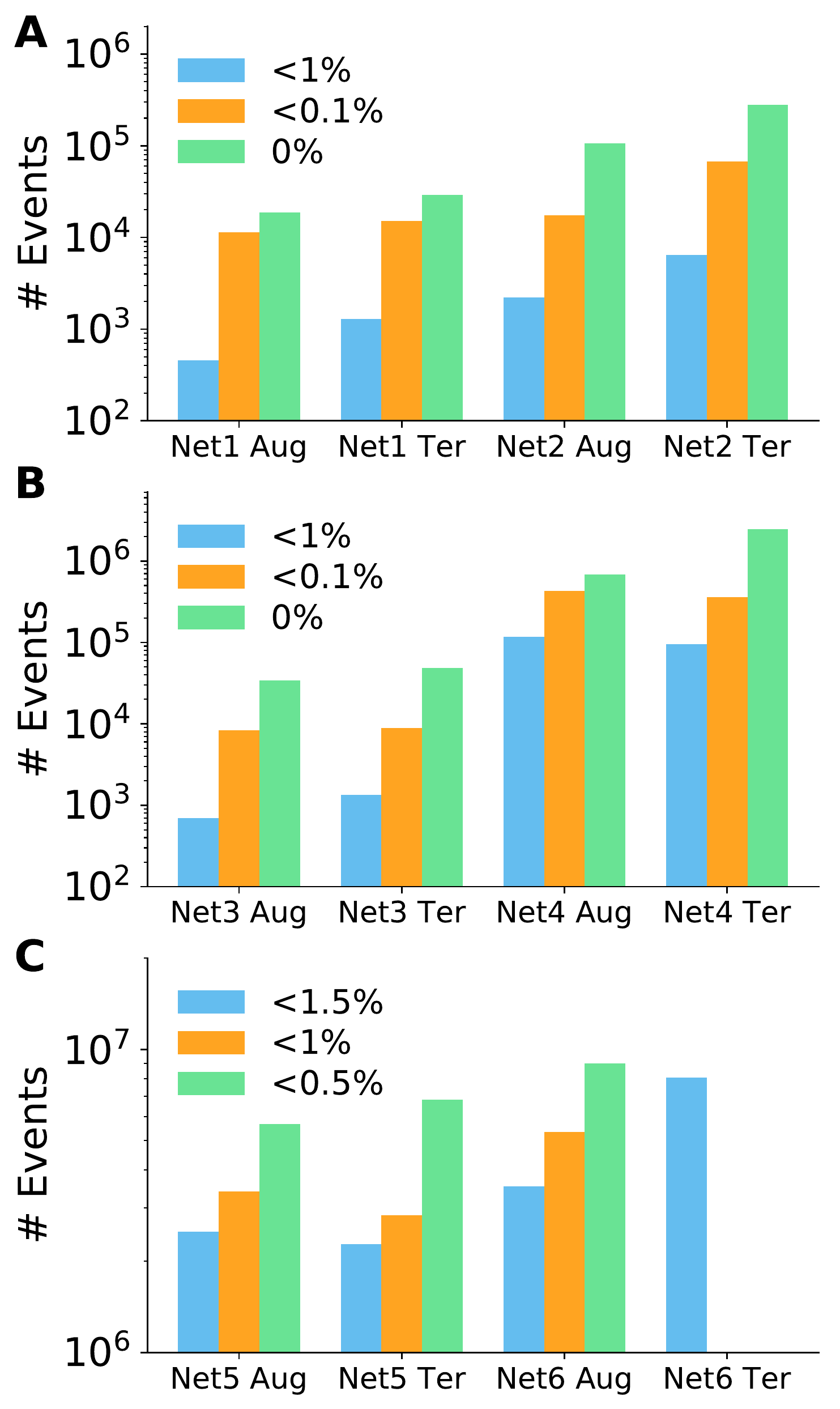}
	\caption{Results of early decision with respect to the required number of events for a certain satisfactory accuracy.}
	\label{early_decision_event}
\end{figure}

Fig.~\ref{early_decision_latency} and Fig.~\ref{early_decision_event} shows the required time steps and the number of events to achieve the acceptable accuracies with both of our methods, respectively.
Note that the results for the TerMapping under several conditions are not shown because it cannot reach a satisfactory accuracy within thousands of time steps.

As can be seen from the figures, with a small compromise on acceptable accuracies, both the number of time steps and spike events required for inference can be significantly reduced. Taking Net2 as an example, a criterion of 1\% accuracy loss for both our AugMapping and TerMapping can lead to the reduction of both time steps and the number of events with around two orders of magnitude as compared to zero-loss conversion.
Although lossless accuracy would be favorable in ANN-to-SNN conversion with deep networks, early decision with a fast speed and small energy consumption could play a more important role in practical applications where efficiency in both time and energy is the main concern.
Notably, the AugMapping still outperforms the TerMapping for almost every case under a given condition. Specifically, the TerMapping requires more time steps and spike events to reach a same accuracy level as the AugMapping. Additionally, for both of our methods, the bigger compromise on accuracy, the more benefit in the efficiency with respect to both time steps and spike events.

Currently, the best reported result for converting a significantly deep network of VGG16 \cite{simonyan2014very} (denoted as Net6 in our study) achieves an accuracy of 91.41\% with a consumption of 793 time steps and $9.342\times10^6$ events \cite{park2019fast}. As is shown in Table~\ref{table3}, our methods result in much higher accuracies as compared to the state-of-the-art one, indicating compromises can be made to further improve efficiency. If we lower the accuracy to 92.76\% which is still significantly better than the current best, our AugMapping only requires $500$ time steps and $7.15\times10^6$ spike events.
Hence, our method outperforms the state-of-the-art with respect to all metrics of accuracy, energy efficiency and speed, highlighting our contribution to improve the spike-based performance.

In order to better quantify the improvements of AugMapping over TerMapping for early decisions, we measure their relative performances on latency and the number of spike events under a same condition.
As is shown in Table~\ref{table4}, AugMapping has a significantly better performance than TerMapping in terms of latency reduction, with a maximum improvement reaching $42.5\times$ for Net3 under the condition of $\textless{}1\%$ accuracy loss. Considering all the other cases, the minimum improvement on the latency is around $3.1\times$.
For the reduction on the number of spike events, AugMapping still outperforms the other one for most cases. Notably, if a perfect accuracy without loss is required, AugMapping achieves a minimum improvement of $1.2\times$ on the reduction of spike events.
The above results thus highlight the advanced performance of AugMapping as is compared to TerMapping.

\begin{table}[!htb]
	\centering
	\caption{Relative performance comparison between AugMapping and TerMapping for early decisions under a certain tolerance on accuracy loss, e.g. `\textless{}1\%'. Relative reductions on latency and the number of spike events are recorded.}
	\begin{tabular}{Sc||Sc Sc Sc|Sc Sc Sc}
		\hline
		\multirow{2}{*}{}
		& \multicolumn{3}{Sc|}{Latency Reduction}         & \multicolumn{3}{Sc}{Spike Events Reduction}                  \\ 
		\cline{2-7}
		& \textless{}1\%   & \textless{}0.1\% & 0\%              & \textless{}1\%   & \textless{}0.1\% & 0\%              \\ \hline
		Net1 & 33${\times}$   & 13.8${\times}$ & 16.3${\times}$ & 2.8${\times}$ & 1.3${\times}$ & 1.5${\times}$ \\
		Net2 & 11${\times}$   & 15.8${\times}$  & 11.5${\times}$ & 2.9${\times}$ & 3.9${\times}$ & 2.7${\times}$ \\
		Net3 & 42.5${\times}$ & 30.9${\times}$   & 40.5${\times}$ & 1.9${\times}$ & 1.1${\times}$ & 1.4${\times}$ \\
		Net4 & 3.1${\times}$  & 3.1${\times}$  & 13.2${\times}$ & 0.8${\times}$ & 0.8${\times}$ & 3.6${\times}$ \\ \hline
		& \textless{}1.5\% & \textless{}1\%   & \textless{}0.5\% & \textless{}1.5\% & \textless{}1\%   & \textless{}0.5\% \\ \hline
		Net5 & 6.3${\times}$  & 6${\times}$    & 8.7${\times}$  & 0.9${\times}$ & 0.8${\times}$ & 1.2${\times}$ \\
		Net6 & 7${\times}$    & -    & -    & 2.3${\times}$ & -   & -   \\ \hline
	\end{tabular}
\label{table4}
\end{table}

\subsection{Evaluation of Approximation}

In this part, we investigate the correlation between the output activations in ANNs and the firing rates in SNNs for both AugMapping and TerMapping, such that the conversion efficacy can be better presented.
We choose Net5 as an example due to its relatively deep structure and the challenge of the task. The firing rates of the output layer are recorded during inference, which are further compared to the output activations of their counterparts in ANN.

In order to quantify the similarity between two vectors such as \textbf{x} and \textbf{y}, we use 
\begin{align}
S(\textbf{x}, \textbf{y})=cos(\theta)=\frac{\textbf{x}\cdot \textbf{y}}{\parallel \textbf{x} \parallel\parallel \textbf{y}\parallel}
\label{eqcos}
\end{align}
where $\theta$ represents the angle between the two vectors. Eq.~(\ref{eqcos}) is used to measure the similarity between the firing rates of SNN and the non-spiking output vectors of ANN.

\begin{figure}[!htb]
	\centering\includegraphics[width=0.48\textwidth]{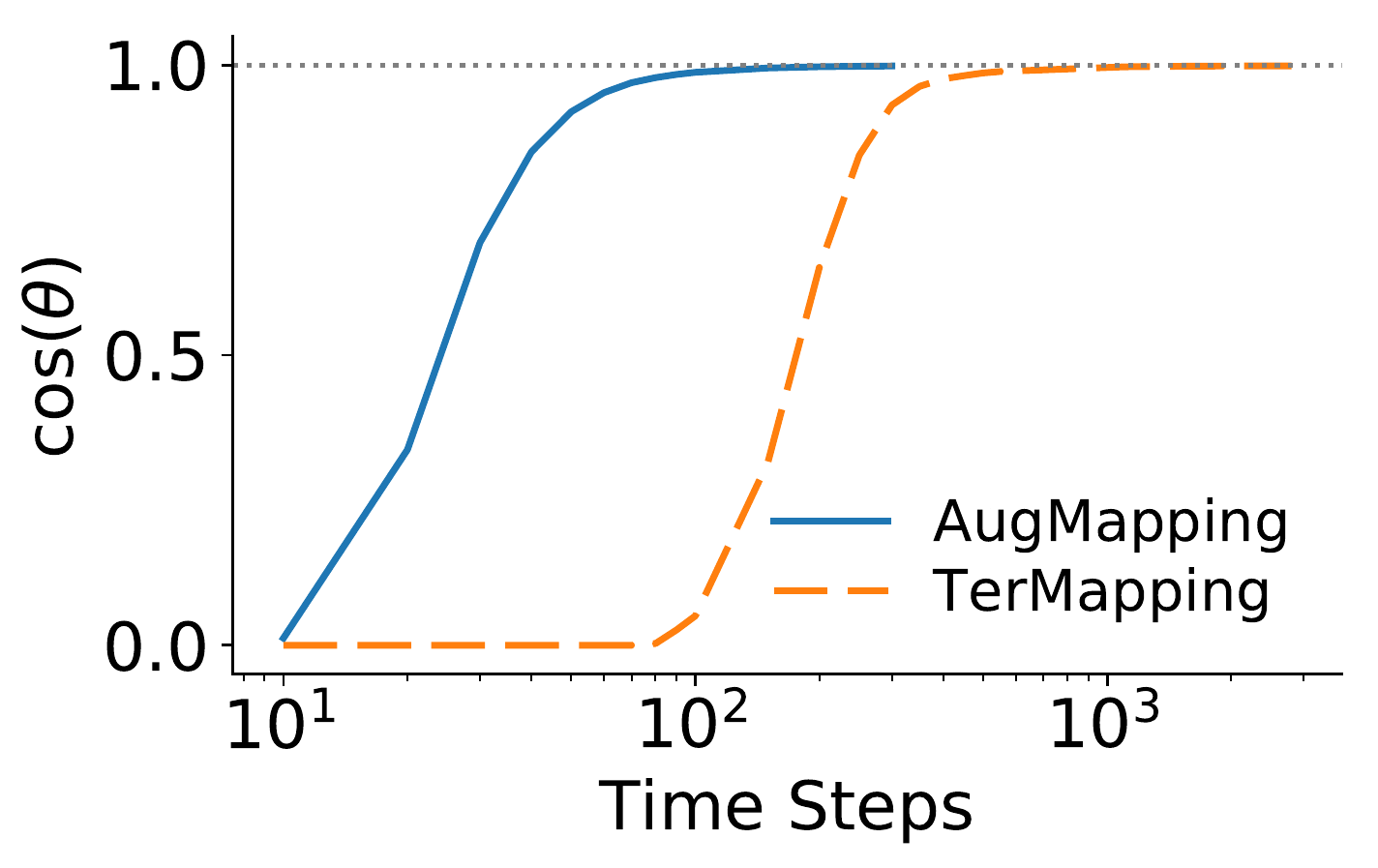}
	\caption{Evaluation of the similarity between the firing rates in SNN and the output activations in ANN for both AugMapping and TerMapping. The horizontal dotted line indicates the perfect match.}
	\label{cos_show}
\end{figure}

Fig.~\ref{cos_show} shows the similarities between the outputs of SNN and ANN during inference for both two of our conversion methods. As can be seen from the figure, both of the two methods can approach a perfect match to the one in ANN with the number of time steps for inference increasing, while the AugMapping is still much faster than the other one.
This reveals the reason underlying the high performance of a conversion scheme. As the latency increases, neurons continuously integrate information such that the firing rates gradually approximate the activations of their counterparts in ANN, thus resulting in a high and nearly lossless accuracy.
This result is also consistent with our theoretical analysis. 

\subsection{Role of Double-Threshold Firing}
\label{SecThreshold}

Our double-threshold scheme is designed to convert both positive and negative activations from ANNs to SNNs. In this part, we will continue to examine its role on recognition performance. The deep Net5 and Net6 on the challenging CIFAR10 benchmark are selected in this experiment.

In order to show the importance of negative activations, we first conduct an experiment on the non-spiking Net5 and Net6 by suppressing all the neurons that are negatively activated. The results show that both Net5 and Net6 suffer a severe decrease, with accuracies down to 35.55\% and 12.36\%, respectively.
This great loss in accuracy suggests that negative outputs in ANNs play an non-trivial role in transferring important information for a remarkable recognition performance.
Therefore, it is important and valuable to represent negative activations in an SNN.

Next, we provide insights into the role of the double-threshold scheme in converted SNNs. Both Net5 and Net6 can be successfully converted into corresponding SNNs with nearly lossless accuracies by either AugMapping or TerMapping. In order to assess the role of our double-threshold scheme, we remove it from the converted SNNs. The SNN results for Net5 only achieve 35.4\% and 32.9\% with the AugMapping and the TerMapping, respectively; the accuracies for converted Net6 are even lower than a chance level.
The significant degradation on accuracy caused by removing the double-threshold firing suggests its importance to realize conversion of lossless accuracy, and highlights its efficacy in representing both positive and negative activations of ANNs.

\subsection{Influence of Boundary Constraint on Augmented Spikes}

Our augmented spikes are capable of addressing the over-activation issue encountered by typical conversion methods \cite{diehl2015fast,rueck2017conversion,sengupta2019going, kim2020spiking} thanks to their advanced form to represent more information with a spike coefficient in a time step.
Here, we examine the effects of spike coefficients by adding a constraint to limit their capability for information representation in a time step. 
We define a parameter $M_\mathrm{aug}$ controlling the maximum number of standard spikes that an augmented one can bundle in one time step.
Therefore, the firing status of our augmented neurons is bounded as:
\begin{align}
o_{i}^{l}(t)=
\begin{cases}
min( M_\mathrm{aug}, o_{i}^{l}(t)), & o_{i}^{l}(t) > 0
\\
max( -M_\mathrm{aug}, o_{i}^{l}(t)), & o_{i}^{l}(t)\leq 0
\end{cases}
\end{align}

\begin{figure}[!htb]
	\centering\includegraphics[width=0.48\textwidth]{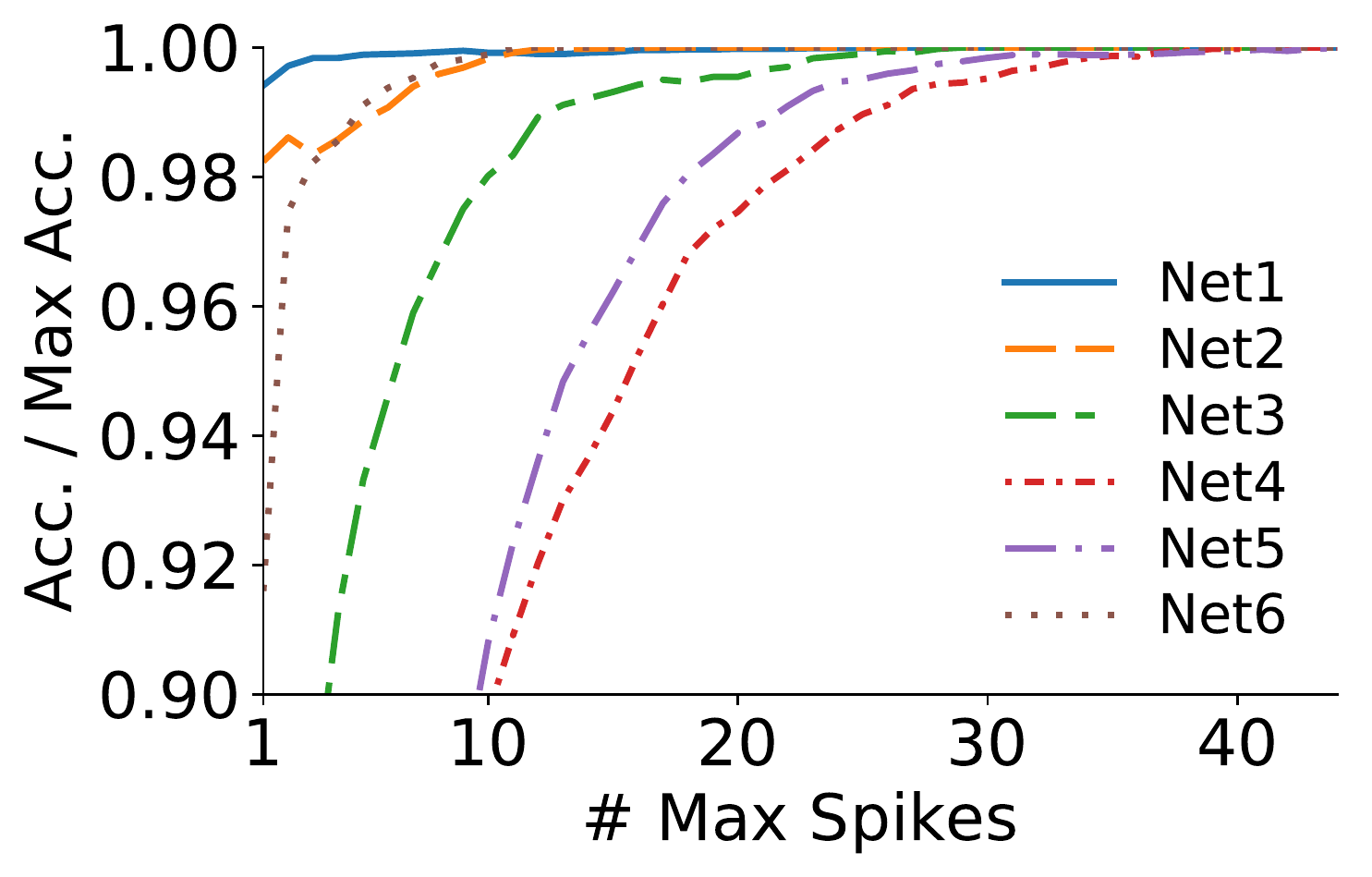}
	\caption{Effects of boundary constraints on augmented spikes. X-axis represents the maximally permitted number of standard spikes packed by one augmented one, while Y-axis denotes the fraction of the accuracies between the constrained and the original networks configured by AugMapping.}
	\label{boundary}
\end{figure}

In our experiment, we first run the converted SNNs without any constraint until their best accuracies are obtained, and the corresponding latency is recorded and then used in the inference with a modified SNN where the boundary constraint on augmented spikes is applied.

Fig.~\ref{boundary} shows the effects of $M_\mathrm{aug}$ on the accuracy performance. When $M_\mathrm{aug}$ is set to 1, augmented spikes are degraded into the typical spikes since each augmented spike can only represent a status of appearance or not in a time step.
As a result, recognition accuracies are decreased for all six networks, and deeper structures suffer much severer loss in accuracy.
As $M_\mathrm{aug}$ increases, accuracies are gradually approaching their corresponding best, since augmented spikes are allowed to carry sufficient information with less constraints.

Notably, as can be seen from Fig.~\ref{boundary}, with a boundary constraint on $M_\mathrm{aug}$ up to 40, our methods can successfully reach a level very close to the cases without constraints. This indicates a few bits are sufficiently enough for a good performance, being beneficial to hardware implementations.


\begin{table*}[!htb]
	\centering
	\caption{Comparison with the other state-of-the-art conversion methods. The digits in parentheses denote the corresponding results with a compromise on accuracy loss.}
	\begin{tabular}{Sc||Sc Sc Sc Sc Sc Sc}
		\hline
		Dataset &
		Method &
		Topology &
		\begin{tabular}[c]{@{}c@{}}ANN \\ acc.\end{tabular} &
		\begin{tabular}[c]{@{}c@{}}SNN \\ acc.\end{tabular} &
		Latency &
		\begin{tabular}[c]{@{}c@{}}\# Events\\ ($10^6$)\end{tabular} \\ \hline
		\multirow{5}{*}{MNIST}  & Weighted Spikes \cite{kim2018deep}     & Net1                             & 98.6\%  & 98.6\%                     & 24                & 8                    \\
		& \textbf{AugMapping} & Net1                             & 98.77\% & \textbf{98.77\% (98.65\%)} & \textbf{46 (23)}  & \textbf{0.02 (0.01)} \\ \cline{2-7} 
		& Weighted Spikes \cite{kim2018deep}      & Net2                             & 99.20\% & 99.20\%                    & 16                & 3                    \\
		& Burst Spikes \cite{park2019fast}        & Net2                             & 99.25\% & 99.25\%                    & 87                & 0.251                \\
		& \textbf{AugMapping} & Net2                             & 99.35\% & \textbf{99.35\% (99.26\%)} & \textbf{65 (12)}  & \textbf{0.11 (0.02)} \\ \hline
		\multirow{7}{*}{CIFAR10} & Cao's Method \cite{cao2015spiking}       & 64c5-p2-64c5-p2-64c3-64-10       & 79.09\% & 77.43\%                    & 400               & 20                   \\
		& Rueckauer's Method \cite{rueck2017conversion} & 32c3-32c3-2s-64c3-64c3-2s-512-10 & 87.86\% & 87.82\%                    & 280               & -                    \\
		& Weighted Spikes \cite{kim2018deep}    & 32c3-32c3-p2-64c3-64c3-p2-512-10 & 89.1\%  & 89.2\%                     & 117               & 400                  \\
		& SpikeNorm \cite{sengupta2019going}          & Net6                             & 91.7\%  & 91.45\%                    & -                 & -                    \\
		& Burst Spikes \cite{park2019fast}        & Net6                             & 91.41\% & 91.1\%                     & 1500              & 49.830               \\
		& \textbf{AugMapping} & Net6                             & 93.42\% & \textbf{93.29\% (92.76\%)}           & \textbf{1100 (500)}     & \textbf{18.10 (7.15)}       \\
		& \textbf{AugMapping} & Net5                             & 94.13\% & \textbf{93.90\% (92.91\%)}   & \textbf{300 (100)} & \textbf{12.51 (3.40)}   \\ \hline
	\end{tabular}
\label{table5}
\end{table*}

\subsection{Comparisons with Other Methods}

In this part, we compare our AugMapping with other state-of-the-art conversion results on MNIST and CIFAR10, in terms of accuracy, latency, and the number of spike events as detailed in Table~\ref{table5}. Fashion-MNIST is not included here as there are few conversion results reported for benchmarking.

The approach of phase-based weighted spikes \cite{kim2018deep} requires a shorter inference latency to reach the lossless accuracy as is compared to ours, but with a cost of significantly larger number of events, indicating a relatively lower energy efficiency.
A recent work with burst spikes \cite{park2019fast} is developed to improve the energy efficiency of converted SNNs by utilizing bursts of binary spikes, yielding a smaller number of events than \cite{kim2018deep} but still larger than ours. More importantly, our latencies outperform the previous works thanks to our augmented scheme where more information can be transmitted with one event at each time step. 


Taking the accuracy as a metric for comparison, our AugMapping has a better performance than all of the other baselines for both MNIST and CIFAR10, regardless of the network type. For instance, under the same condition with the large Net6 structure and the more challenging CIFAR10 dataset, our AugMapping obtains an accuracy of 93.29\% that is significantly better than the other methods. With a different network variant of Net5, the accuracy can be further improved. The remarkable performance of our methods can be attributed to both double-threshold firing and augmented spikes, enabling spiking neurons to represent both positive and negative activations with more informative spikes, such that more advanced techniques in ANNs can be adopted for improvements.
Notably, both the latency and energy efficiency of our AugMapping can be further improved with a compromise on acceptable accuracies.

In summary, our AugMapping is faster, more accurate and energy-efficient as is compared to the other state-of-the-art baselines, making it of great merit for applied implementations of deep SNNs.

\section{Discussions}
\label{SecDis}

Despite of the advatanges of SNNs such as energy efficiency \cite{merolla2014million,davies2018loihi,roy2019towards}, their accuracies resulted from direct training mechanisms still lag far behind those of ANNs \cite{pfeiffer2018deep,tavanaei2019deep,taherkhani2020a}. Differently, converting a pre-trained ANN to an SNN provides a straightforward and yet effective mechanism to narrow the accuracy gap between SNNs and ANNs \cite{perez2013mapping,cao2015spiking}. The accuracy of the converted SNN can be improved by either enhancing the one of its corresponding ANN or by reducing performance loss due to conversion. A standard SNN typically has a single firing threshold, being limited to represent positive activations only. As a result, the advanced activation functions like LeakReLU in ANNs \cite{xu2015empirical, maas2013rectifier} cannot be fully exploited. In order to overcome this challenge, we introduce a double-threshold firing scheme, where both positive and negative thresholds are used for the neuron to elicit polarized spikes. In return, our methods can take advantages of superior ANNs to achieve better performance.

Our double-threshold firing scheme is firstly used to extend a typical threshold-balancing method of DataNorm \cite{diehl2015fast}, and thus TerMapping is developed. The TerMapping inherits the advantages of DataNorm and can successfully achieve nearly lossless conversion (see Table~\ref{table3}). Differently, TerMapping is more accurate than DataNorm thanks to the double-threshold firing scheme providing approaches for mapping both positive and negative activations (see Table~\ref{table2}). However, drawbacks from DataNorm are also brought to TerMapping, such as complicated procedures for configuring proper network parameters and inefficiency in both time and energy (see Fig.~\ref{cmp_speed} and Fig.~\ref{cmp_events}). Notably, our double-threshold scheme can be easily generalized to other conversion-based methods, and it could be applied to a broad range of ANNs with both positive and negative activations.

In order to further improve the efficiency of the conversion approach, we introduce a new scheme of augmented spike that employs spike coefficients to carry the number of typical all-or-nothing spikes occurring at a time step. Based on this, a new conversion method called AugMapping is developed 
with a clear and simple rule to assign network parameters for SNNs.
Importantly, our AugMapping can not only achieve nearly lossless conversion but also consumes significantly smaller number of time steps and spike events as compared to TerMapping (see Table~\ref{table3} and Table~\ref{table4}), highlighting its advantage of high efficiency in energy. 
Additionally, the augmented scheme endows spikes with the advanced ability to carry more information at one time step, thus significantly reducing the latency for information accumulation. As a result, our AugMapping is also efficient in time as a low latency is required. Notably, our augmented scheme could be also favorable for hardware implementations since a few bits could be sufficiently enough for a high performance (see Fig.~\ref{boundary}).

Three realistic datasets with various networks are used to investigate the effectiveness of our methods. The better performance of our methods over the state-of-the-art baselines (see Table~\ref{table5}) highlight the potential merit of our approaches. In addition to the image recognition tasks, our methods could be easily generalized to other challenging problems such as object detection, for which our advantageous performance could be favorable and beneficial.

\section{Conclusion}
\label{secCon}

In this work, we first introduced a double-threshold scheme for SNNs to fully benefit from advanced ANNs in the ANN-to-SNN conversion. Then, we developed TerMapping by extending the DataNorm method with our double-threshold scheme.
Moreover, another new scheme of augmented spikes is introduced to represent more information at one time step. Accordingly, a new AugMapping was developed for ANN-to-SNN conversion, but importantly with a simple and clear rule to configure spiking neurons in contrast to the complicated threshold-balancing approaches in other related works. We investigated the performance of our methods with various networks based on three challenging datasets. Experimental results show that our double-threshold scheme benefits the improvement of accuracy for SNNs. Moreover, our advanced AugMapping is more advantageous for constructing accurate, fast and efficient deep SNNs than the state-of-the-art baselines, which could be greatly valuable for neuromorphic computing.


\ifCLASSOPTIONcaptionsoff
  \newpage
\fi

\begin{IEEEbiography}[{\includegraphics[width=1in,height=1.25in,clip,keepaspectratio]{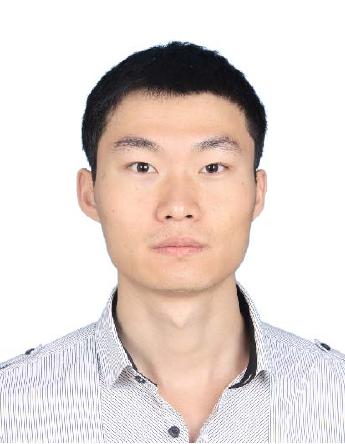}}]{Qiang~Yu}
	(M'12) received the B.Eng. degree in electrical engineering and automation from the Harbin Institute of Technology, Harbin, China, in 2010, and the Ph.D. degree in electrical and computer engineering from the National University of Singapore, Singapore, in 2014.\\
	He is an Associate Professor with the College of Intelligence and Computing,
	Tianjin University, Tianjin, China. Before that, he was a Post-Doctoral Research Fellow with the Max-Planck-Institute for Experimental Medicine, G\"{o}ttingen, Germany, from 2014 to 2016, and a Research Scientist in the Institute for Infocomm Research, Agency for Science, Technology and Research, Singapore, from 2016. He is a recipient of the 2016 IEEE Outstanding TNNLS Paper Award. His current research interests include learning algorithms in spiking neural networks, neural coding, cognitive computations and machine learning.
\end{IEEEbiography}

\begin{IEEEbiography}[{\includegraphics[width=1in,height=1.25in,clip,keepaspectratio]{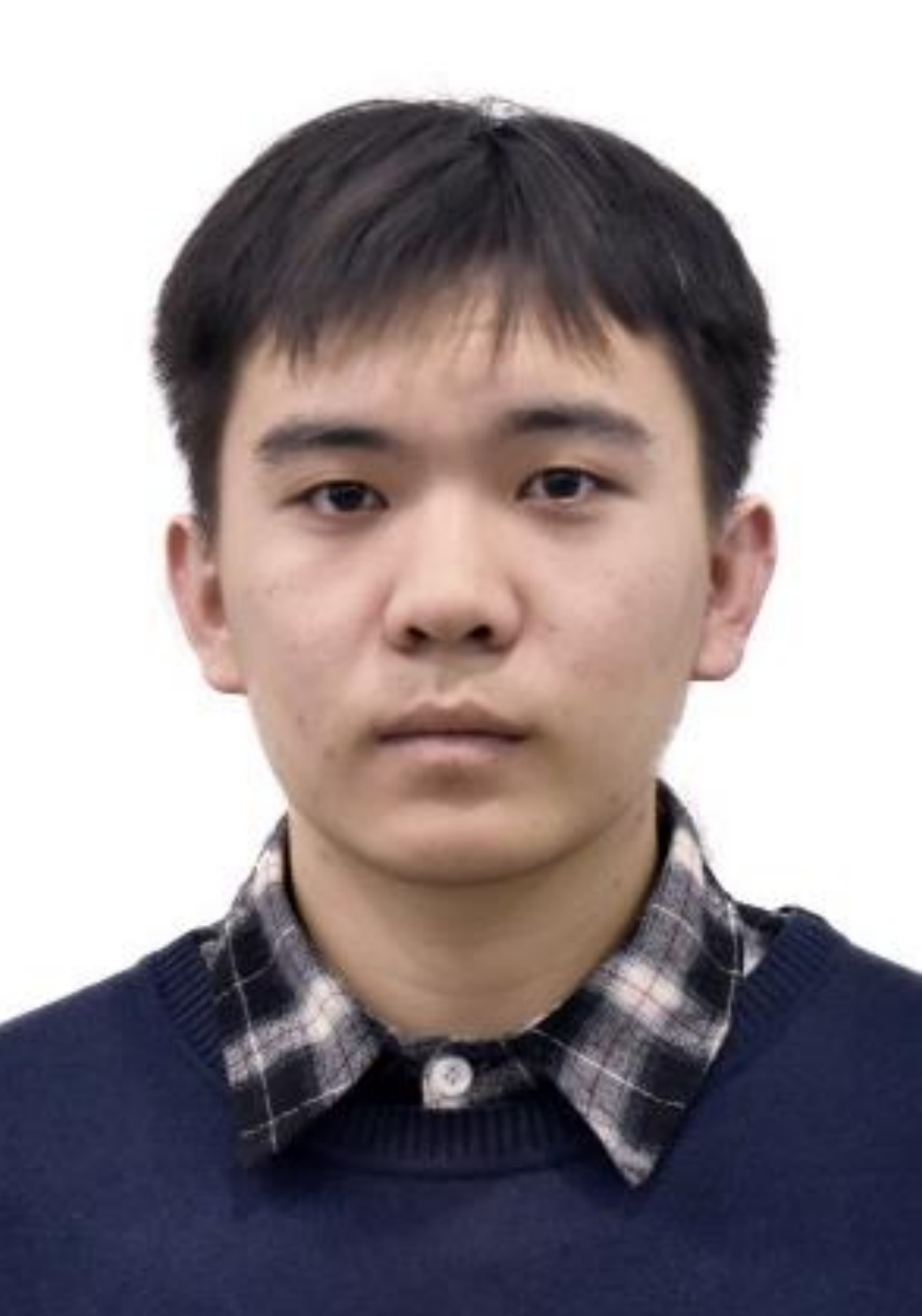}}]{Chenxiang~Ma}
	received the B.Eng. degree from the China University of Petroleum, Qingdao, China, in 2019. He is currently pursuing the master's degree with the College of Intelligence and Computing, Tianjin University, Tianjin, China. His current research interests include learning algorithms in spiking neural network and deep learning.
\end{IEEEbiography}

\begin{IEEEbiography}[{\includegraphics[width=1in,height=1.25in,clip,keepaspectratio]{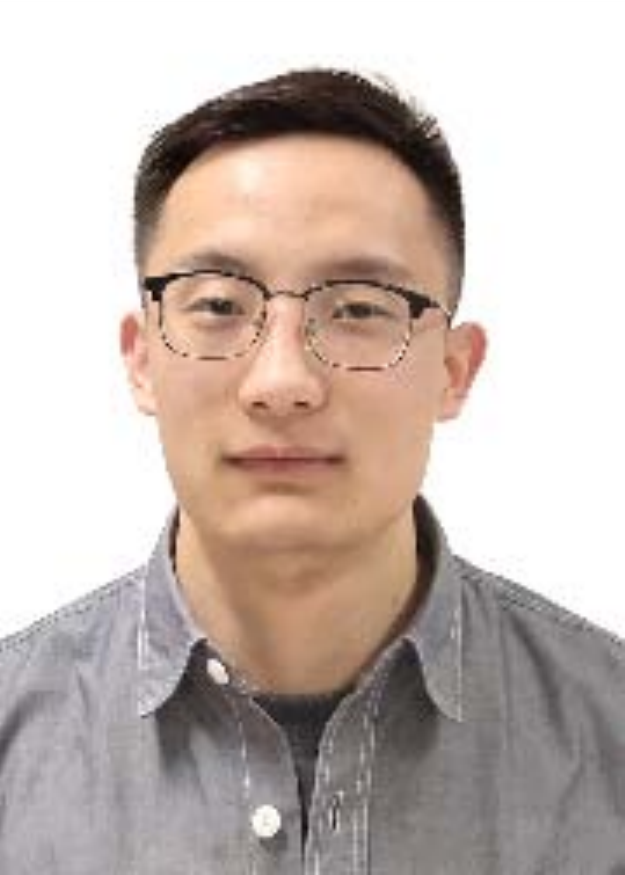}}]{Shiming~Song}
	received the bachelor's degree from Sichuan University, Chengdu, China, in 2018. He is currently pursuing the master's degree with the College of Intelligence and Computing, Tianjin University, Tianjin, China. His current research interests include spike-based learning, neural encoding and machine learning.
\end{IEEEbiography}

\begin{IEEEbiography}[{\includegraphics[width=1in,height=1.25in,clip,keepaspectratio]{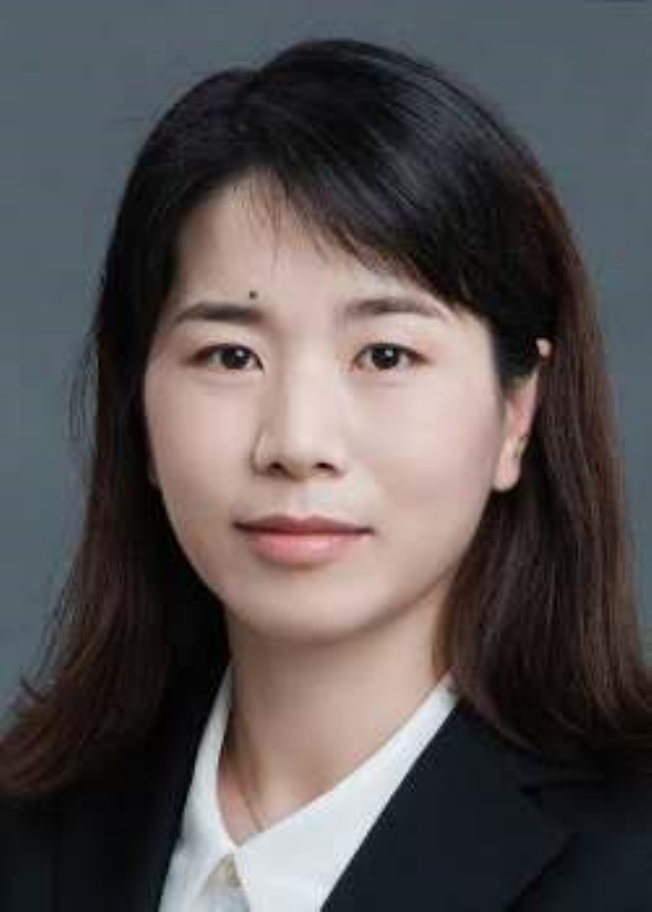}}]{Gaoyan~Zhang}
	received the B.S. degree in communication engineering from Henan Normal University, Xinxiang, China, in 2008 and the Ph.D. degree in cognitive neuroscience from Beijing Normal University, Beijing, China in 2014. 
	She is an Associate Professor with College of Intelligence and Computing, Tianjin University, Tianjin, China since 2014. From 2017 to 2018, she was a visiting scholar in National Institutes of Health, USA. Her research interest includes brain signal and information processing, especially fMRI and EEG data analysis and modeling.	
\end{IEEEbiography}

\begin{IEEEbiography}[{\includegraphics[width=1in,height=1.25in,clip,keepaspectratio]{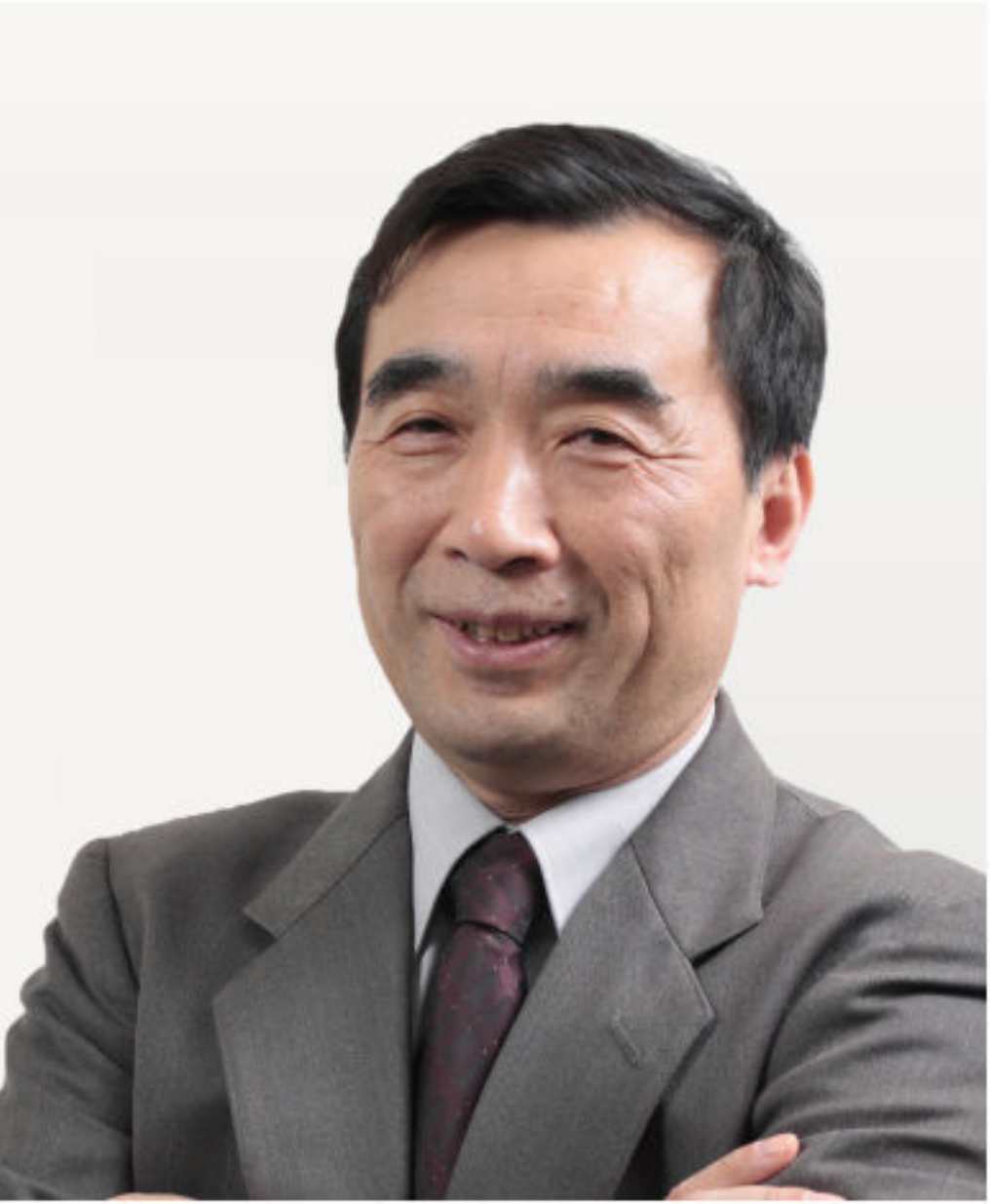}}]{Jianwu~Dang}
	(M'12) graduated from Tsinghua Univ., China, in 1982, and got his M.S. degree at the same university in 1984. He worked for Tianjin Univ. as a lecture from 1984 to 1988. He was awarded the PhD degree from Shizuoka Univ., Japan in 1992. He worked for ATR Human Information Processing Labs., Japan, as a senior researcher from 1992 to 2001. He joined the University of Waterloo, Canada, as a visiting scholar for one year from 1998. Since 2001, he has worked for Japan Advanced Institute of Science and Technology (JAIST) as a professor. He joined the Institute of Communication Parlee (ICP), Center of National Research Scientific, France, as a research scientist the first class from 2002 to 2003. Since 2009, he has joined Tianjin University, Tianjin, China. His research interests are in all the fields of speech science including brain science, and speech signal processing. He built MRI-based bio-physiological models for speech and swallowing, and endeavors to apply these models on clinics.
\end{IEEEbiography}

\begin{IEEEbiography}[{\includegraphics[width=1in,height=1.25in,clip,keepaspectratio]{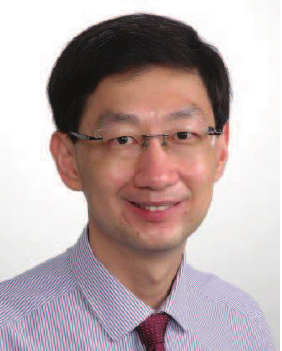}}]{Kay~Chen~TAN}
	(SM'08-F'14) received the B.Eng. (First Class Hons.) degree in electronics and	electrical engineering and the Ph.D. degree from the University of Glasgow, Glasgow, U.K., in 1994 and 1997, respectively.\\
	He is a Full Professor with the Department of Computer Science, City University of Hong Kong, Hong  Kong. He has published over 200 refereed
	articles and five books.\\
	Dr.	Tan	is the Editor-in-Chief of the IEEE TRANSACTIONS	ON EVOLUTIONARY
	COMPUTATION, was the Editor-in-Chief of the	IEEE Computational Intelligence Magazine from 2010 to 2013, and currently serves as the Editorial Board Member of over 20 journals. He is an elected member of
	the IEEE CIS AdCom from 2017 to 2019 and is an IEEE CIS Distinguished
	Lecturer from 2015 to 2017.
\end{IEEEbiography}

\end{document}